\documentclass[10pt,twocolumn,letterpaper]{article}

\usepackage{cvpr}
\usepackage{times}
\usepackage{epsfig}
\usepackage{graphicx}
\usepackage{amsmath}
\usepackage{amssymb}
\usepackage{arydshln}
\usepackage[hyphens]{url}
\usepackage{multirow}
\usepackage{enumitem}

\newcommand{\argmin}{\operatornamewithlimits{argmin}}

\usepackage[breaklinks=true,colorlinks,bookmarks=false]{hyperref}

\cvprfinalcopy 

\def\cvprPaperID{****} 

\ifcvprfinal\fi
\begin{document}

\title{Robust Point Cloud Based Reconstruction of Large-Scale Outdoor Scenes}

\author{Ziquan Lan \qquad Zi Jian Yew \qquad Gim Hee Lee\\
Department of Computer Science, National University of Singapore\\
{\tt\small \{ziquan, zijian.yew, gimhee.lee\}@comp.nus.edu.sg}
}

\maketitle

\begin{abstract}
Outlier feature matches and loop-closures that survived front-end data association can lead to catastrophic failures in the back-end optimization of large-scale point cloud based 3D reconstruction. To alleviate this problem, we propose a probabilistic approach for robust back-end optimization in the presence of outliers. More specifically, we model the problem as a Bayesian network and solve it using the Expectation-Maximization algorithm.
Our approach leverages on a long-tail Cauchy distribution to suppress outlier feature matches in the odometry constraints, and a Cauchy-Uniform mixture model with a set of binary latent variables to simultaneously suppress outlier loop-closure constraints and outlier feature matches in the inlier loop-closure constraints.
Furthermore, we show that by using a Gaussian-Uniform mixture model, our approach degenerates to the formulation of a state-of-the-art approach for robust indoor reconstruction. Experimental results demonstrate that our approach has comparable performance with the state-of-the-art on a benchmark indoor dataset, and outperforms it on a large-scale outdoor dataset. Our source code can be found on the project website \url{https://github.com/ziquan111/RobustPCLReconstruction}.
\vspace{-8pt}
\end{abstract}

\section{Introduction}

\begin{figure}
    \centering
    \includegraphics[width=0.47\textwidth]{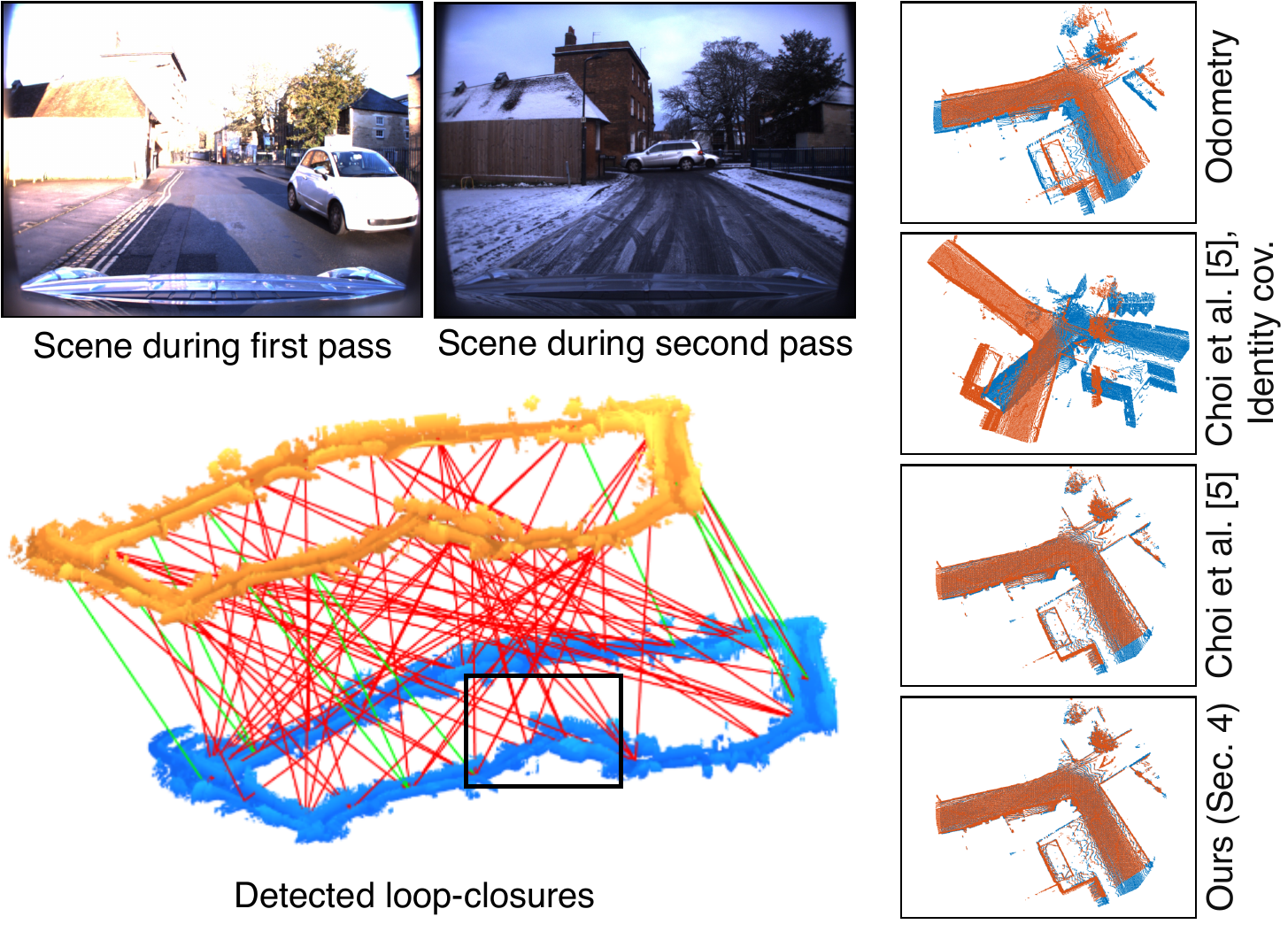}
    \caption{Reconstruction of a 1km route traversed in two different seasons: summer (orange) and winter (blue). 
    The outlier (red links) loop-closures significantly outnumber the inliers (green links). Four zoomed-in point clouds on the right are reconstructed from different methods. }
    \label{fig:short-link}
\end{figure}

Point cloud reconstruction of outdoor scenes has many important applications such as 3D architectural modeling, terrestrial surveying, Simultaneous Localization and Mapping (SLAM) for autonomous vehicles, \etc.
Compared to images, point clouds from 3D scanners exhibit less variation under different weather or lighting conditions, \eg, summer and winter (Fig. \ref{fig:short-link}), or day and night (Fig. \ref{fig:long-link}).  
Furthermore, the depths of point clouds from 3D scanners are more accurate than image-based reconstructions. Consequently, point clouds from 3D scanners are preferred for large-scale outdoor 3D reconstructions.
Most existing methods for 3D reconstruction are solved via a two-step approach: a front-end data association step and a back-end optimization step. 
More specifically, data association is used to establish feature matches \cite{zijian} in point cloud fragments for registration, and loop-closures \cite{mika} between point cloud fragments for pose-graph \cite{olson2006fast} optimization. Unfortunately, no existing algorithm for feature matching and loop-closure detection guarantees complete elimination of outliers.
Although outlier feature matches are usually handled with RANSAC-based geometric verification \cite{Lee14GeoVerification,zijian}, such pairwise checks do not consider global consistency. 
In addition, the numerous efforts on improving the accuracy in loop-closure detection \cite{cummins2011appearance, bagofwords, vocabularytree, mika} are not completely free from false positives.
Many back-end optimization algorithms \cite{factorgraph, g2o, olson2006fast} are based on non-linear least-squares that lack the robustness to cope with outliers. A small number of outliers would consequently lead to catastrophic failures in the 3D reconstructions. 
Several prior works focus on disabling outlier loop-closures in the back-end optimization \cite{baseline, em, switchable}. However, these methods do not consider the effect from the outlier feature matches with the exception of \cite{zhou2016fast} that solves global geometric registration in a very small-scale problem setting. 

The main contribution of this paper is a probabilistic approach for robust back-end optimization to handle outliers from a weak front-end data association
in large-scale point cloud based reconstructions.
Our approach simultaneously suppresses outlier feature matches and loop-closures. 
To this end, we model our robust point cloud reconstruction problem as a Bayesian network. The global poses of the point cloud fragments are the unknown parameters, and odometry and loop-closure constraints are the observed variables. A binary latent variable is assigned to each loop-closure constraint; it determines whether a loop-closure constraint is an inlier or outlier.
We model feature matches in the odometry constraints with a long-tail Cauchy distribution to gain robustness to outlier matches. Additionally, we use a Cauchy-Uniform mixture model for loop-closure constraints. The uniform and Cauchy distributions model outlier loop-closures and the feature matches in inlier loop-closures, respectively. In contrast to many existing back-end optimizers that use rigid transformations as the odometry and loop-closure constraints \cite{baseline, g2o, em, olson2006fast, switchable}, we use the distances between feature matches to exert direct influence on these matches. 

We use the Expectation-Maximization (EM) algorithm \cite{celeux1992classification,em} to find the globally consistent poses of the point cloud fragments (Sec. \ref{sec:EM}). The EM algorithm iterates between the Expectation and Maximization steps. In the Expectation step, the posterior of a loop-closure constraint being an inlier is updated. In the Maximization step, a local optimal solution for the global poses is found from maximizing the expected complete data log-likelihood over the posterior from the expectation step.
We also generalize our approach to solve reconstruction problems with an easier setting  (Sec. \ref{sec:generalization}). In particular, a strong assumption is imposed: odometry and inlier loop-closure constraints are free from outlier feature matches. We show that by using a Gaussian-Uniform mixture model, our approach degenerates to the formulation of a state-of-the-art approach for robust indoor reconstruction \cite{baseline}. Fig. \ref{fig:short-link} shows an example of the reconstruction result with our method compared to other methods in the presence of outliers.

\section{Related Work}
Reconstruction of outdoor scenes has been studied in \cite{related_rgb, related_outdoor_noloop1}. Sch{\"o}ps et al. \cite{related_outdoor_noloop1} propose a set of filtering steps to detect and discard unreliable depth measurements acquired from a RGB-D camera. However, loop-closures is not detected and this can lead to reconstruction failures. Relying on very accurate GPS/INS, Pollefeys et al. \cite{related_rgb} propose a 3D reconstruction system from RGB images. However, GPS/INS signal may be unavailable or unreliable, especially on cloudy days or in urban canyons. Our work relies on neither GPS/INS nor RGB images. In contrast, we focus on reconstruction from point cloud data acquired from 3D scanners that is less sensitive to weather or lighting changes.
There are also many works on indoor scene reconstruction. 
Since the seminal KinectFusion \cite{kinectfusion}, there are several follow-up algorithms \cite{related_noloop2, related_noloop3, kintinuous}. Unfortunately, these methods do not detect loop-closures. Nonetheless, there are many RGB-D reconstruction methods with loop-closure detection \cite{baseline, related_rgbd1, related_rgbd2, related_rgbd3, related_rgbd4, related_rgbd5, related_rgbd6, related_rgbd7, related_rgbd8}. 

Choi et al. \cite{baseline} achieve the state-of-the-art performance for indoor reconstruction with robust loop-closure. 
However, they assume no outlier feature matches in the odometry and inlier loop-closure constraints. We relax this assumption to achieve robust feature matching.
More specifically, \cite{baseline} estimates a switch variable \cite{switchable} for each loop-closure constraint using line processes \cite{lineprocess}. Outlier loop-closures are disabled by setting the respective switch variables to zero. Additional switch prior terms are imposed and chosen empirically \cite{switchable}
to prevent a trivial solution of removing all loop-closure constraints. In comparison, our approach does not require the additional prior terms. We estimate the posterior of a loop-closure being an inlier constraint in the Expectation step shown in Sec. \ref{sec:EM}. 
The EM approach is also used by Lee et al. \cite{em}. However, they solve a robust pose-graph optimization problem without coping with the feature matches for reconstruction.

\section{Overview}\label{sec:overview}
In this section, we provide an overview of our reconstruction pipeline that consists of four main components: point cloud fragment construction, point cloud registration, loop-closure detection, and robust reconstruction with EM.

\paragraph{Point cloud fragment construction.}
A single scan from a 3D scanner, \eg LiDAR, contains limited number of points. 
We integrate multiple consecutive scans with odometry readings obtained from dead reckoning \eg, the Inertial Navigation System (INS) \cite{oxford} to form local point cloud fragments. A set of 3D features is then extracted from each point cloud fragment using \cite{zijian}. 

\paragraph{Point cloud registration.}

The top $k_1$ feature matches between two consecutive point cloud fragments $F_i$ and $F_{i+1}$ are retained as the odometry constraint $X_{i,i+1}$.  
Since consecutive fragments overlap sufficiently by construction \cite{oxford, zijian}, we define $X_{i,i+1}$ as a reliable constraint but note that it can contain outlier feature matches. 

\paragraph{Loop-closure detection.}
It is inefficient to perform an exhaustive pairwise registration
for large-scale outdoor scenes with many point cloud fragments. 
Hence, we perform point cloud based place-recognition \cite{mika} to identify a set of candidate loop-closures.
We retain the top $k_2$ potential loop-closures for each fragment and remove the duplicates. For each loop-closure between fragments $F_i$ and $F_j$, we keep the set of top $k_1$ feature matches denoted as $Y_{ij}$. We define $Y_{ij}$ as a loop-closure constraint, which can either be an inlier or outlier. Similar to the odometry constraint, an inlier loop-closure $Y_{ij}$ can also contain outlier feature matches.

\paragraph{Robust reconstruction with EM.}
The constraints from point cloud registration and loop-closure detection can contain outliers. In particular, both odometry and loop-closure constraints can contain outlier feature matches. Moreover, many detected loop-closures are false positives. In the next section, we describe our probabilistic modeling approach to simultaneously suppress outlier feature matches and false loop-closures. The EM algorithm is used to solve for the globally consistent fragment poses. Optional refinement using ICP can be applied to further improve the global point cloud registration.

\section{Robust Reconstruction with EM} \label{sec:EM}
\begin{figure}
    \centering
    \includegraphics[width=0.35\textwidth]{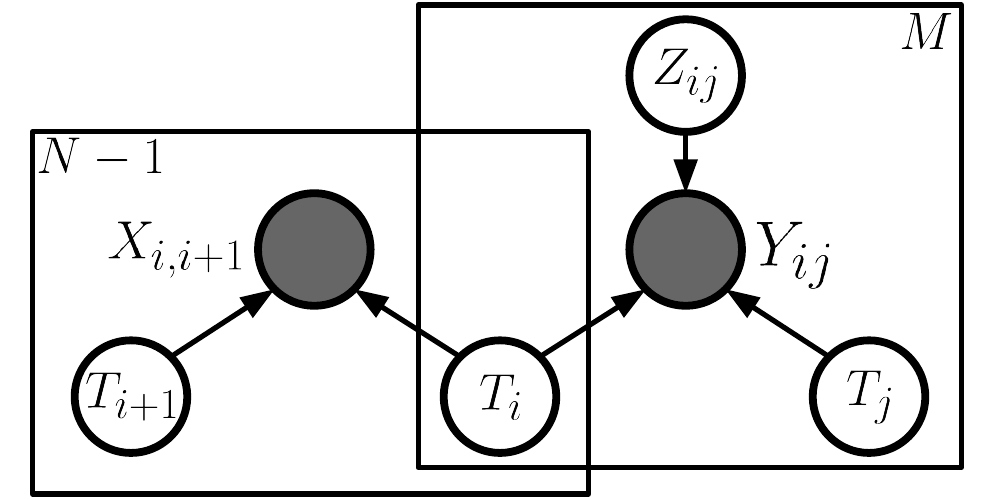}
    \caption{Bayesian network representation of the robust reconstruction problem. $T_{i+1}$, $T_i$ and $T_j$ are fragment poses. $X_{i,i+1}$ is an odometry constraint. $Y_{ij}$ is a loop-closure constraint. $Z_{ij}$ is an assignment variable. $N-1$ and $M$ indicate the numbers of odometry constraints and loop-closure constraints respectively.}
    \label{fig:model}
\end{figure}

We model the robust reconstruction problem as a Bayesian network shown in Fig.~\ref{fig:model}. Let $T =[T_1, ..., T_i, ..., T_N]^\top$, where $T_i \in \text{SE}(3)$, denote the $N$ fragment poses, $X = [X_{12}, ..., X_{i,i+1}, ..., X_{N-1,N}]^\top$ denote the $N-1$ odometry constraints obtained in point cloud registration, and $Y = [...,Y_{ij},...]^\top$ denote the $M$ loop-closure constraints obtained in loop-closure detection. We explicitly assign the loop-closure constraints into $2$ clusters that represent the inliers and outliers. For each loop-closure constraint $Y_{ij}$, we introduce a corresponding assignment variable $Z_{ij} = [Z_{ij,in}, Z_{ij,out}]^\top \in \{[1,0]^\top, [0,1]^\top\}$. $Z_{ij}$ is a one-hot vector: $Z_{ij,in}=1$ and $Z_{ij,out}=1$ assigns assigns $Y_{ij}$ as an inlier and outlier loop-closure constraint, respectively. 
We use $Z = [...,Z_{ij},...]^\top$ to denote the assignment variables. $T$ is the unknown parameter, $Z$ is the latent variable, and $X$ and $Y$ are both observed variables. 

Robust reconstruction can be solved as finding the Maximum a Posterior (MAP) solution of $p(T \vert X, Y)$. However, the MAP solution involves an intractable step of marginalization over the latent variable $Z$. We circumvent this problem by using the EM algorithm that takes the 
maximization of the expected complete data log-likelihood over the posterior of the latent variables. 
The EM algorithm iterates between the Expectation and Maximization steps.
In the Expectation step, we use $T^{old}$, \ie, fragment poses solved from the previous iteration to find the posterior distribution of the latent variable $Z$,
\begin{equation}\label{eq:z|ythetaold}
    p(Z \vert Y,T^{old})
    = \frac{p(Y \vert Z,T^{old}) p(Z \vert T^{old})}{p(Y \vert T^{old})},
\end{equation}
in which $Z$ does not depend on $X$, since they are conditionally independent given $Y$ according to the Bayesian network in Fig. \ref{fig:model}.

In the Maximization step, the posterior distribution (Eq. (\ref{eq:z|ythetaold})) is used to update $T$ by maximizing the expectation of the complete data log-likelihood denoted by
\begin{align}\label{eq:q}
    Q_{EM} :&= \sum_{Z} p(Z \vert Y, T^{old}) \ln{p(X,Y,Z \vert T)} \\
    &= \underbrace{\ln{p(X \vert T)}}_{Q_X} + 
    \underbrace{\sum_{Z} p(Z \vert Y, T^{old}) \ln{p(Y, Z \vert T)}}_{Q_Y}. \notag
\end{align}
We define $Q_X$ for the term with odometry constraints,
and $Q_Y$ for the term with loop-closure constraints.

\paragraph{Initialization.} The unknown parameters, \ie, global poses $T$ of the $N$ fragments, are initialized with the relative poses computed from odometry constraints $X$ using ICP. Other dead reckoning methods such as wheel odometry and/or INS readings can also be used.

\subsection{Modeling Odometry Constraints}
Odometry constraints are obtained from point cloud registration between two consecutive point cloud fragments. Recall that an odometry constraint $X_{i,i+1}$ is a set of feature matches between fragments $F_i$ and $F_{i+1}$, which can contain outlier matches. To gain robustness, we model each feature match $(\mathbf{p}, \mathbf{q}) \in X_{i,i+1}$ with a long-tail multivariate Cauchy distribution. Suppose these feature matches are independent and identically distributed (i.i.d.), we take a geometric mean over their product to get
\begin{equation}\label{eq:xii+1}
    p(X_{i,i+1} \vert T)= \Big( \prod_{
    (\mathbf{p}, \mathbf{q}) \in X_{i,i+1}} 
    \textit{Cauchy}_{i,i+1}(\mathbf{p},\mathbf{q}) \Big)^\frac{1}{\vert X_{i,i+1} \vert},
\end{equation}
where
\begin{equation}
    \textit{Cauchy}_{i,i+1}(\mathbf{p},\mathbf{q}) =  \frac{1}{\pi^2 \sqrt{\det\Sigma} (1 + d_{\Sigma}^2(T_i \mathbf{p}, T_{i+1} \mathbf{q}) )^2},
\end{equation}
which we assume an isotropic covariance $\Sigma = \sigma^2 \mathbf{I}$ with scale $\sigma$, and $d_{\Sigma}$ denotes the Mahalanobis distance such that
\begin{equation}\label{eq:dsigma}
    d_{\Sigma}^2(T_i \mathbf{p}, T_{i+1} \mathbf{q}) = (T_i \mathbf{p} - T_{i+1} \mathbf{q})^\top \Sigma^{-1} (T_i \mathbf{p} - T_{i+1} \mathbf{q}).
\end{equation}
The value of $\sigma$ is set based on the density of extracted features. For example, $\sigma = 0.5$m in the outdoor dataset.





\subsection{Modeling Loop-Closure Constraints}
A loop-closure constraint $Y_{ij}$ is the set of feature matches between fragments $F_i$ and $F_j$. We propose to use a Cauchy-Uniform mixture model to cope with the (1) outlier loop-closure constraints and (2) outlier feature matches in the inlier loop-closure constraints.

To distinguish between inlier and outlier loop-closures, we model the distribution of assignment variable $Z$ as a Bernoulli distribution defined by the inlier probability $\lambda \in [0,1]$,
\begin{equation} \label{eq:zij}
    p(Z_{ij}) = \lambda^{Z_{ij,in}} (1-\lambda)^{Z_{ij,out}}.
\end{equation}

Next, we use two distributions: Cauchy and Uniform distributions to model the inlier and outlier loop-closure constraints, respectively.
\paragraph{Cauchy distribution -- inlier loop-closure constraints.}
The inlier loop-closure constraints can contain outlier feature matches. We use the same multivariate Cauchy distribution as Eq. (\ref{eq:xii+1}) and further reorganize the terms. We define $C_{ij}$ for brevity, such that
\begin{equation}\label{eq:cij}
    C_{ij} := p(Y_{ij} \vert Z_{ij,in} = 1, T) = \pi^{-2} \sigma^{-3} e^{-2A_{ij}},
\end{equation}
in which
\begin{equation}\label{eq:aij}
    A_{ij} = \frac{1}{\vert Y_{ij} \vert} \sum_{(\mathbf{p}, \mathbf{q}) \in Y_{ij}} {\ln (1 + \frac{\Vert T_i \mathbf{p} - T_{j} \mathbf{q} \Vert^2}{\sigma^2})},
\end{equation}
and $\vert Y_{ij}\vert$ denotes the number of feature matches in $Y_{ij}$.


\paragraph{Uniform distribution -- outlier loop-closure constraints.} 
We model the outlier loop-closure constraints with a uniform distribution defined by a constant probability $u \in (0, 1)$,
\begin{equation}\label{eq:yij|zout}
    p(Y_{ij} \vert Z_{ij,out} = 1, T) = u.
\end{equation}

\subsection{Expectation Step}
Recall that the expectation step is evaluated in Eq. (\ref{eq:z|ythetaold}). 
Plugging Eq. (\ref{eq:zij}), (\ref{eq:cij}) and (\ref{eq:yij|zout}) into the Bayes' formula, we obtain the posterior of being an inlier loop-closure constraint,
\begin{align}\label{eq:lijin}
    \mathcal{P}_{ij}^{in} &:= p(Z_{ij,in} = 1 \vert Y, T) = \frac{\Theta}{\Theta + e^{2A_{ij}}},
\end{align}
where
\begin{equation}
    \Theta = \frac{\lambda}{(1-\lambda)u \pi^2 \sigma^3}.
\end{equation}
The constant $\Theta$ consists of two distribution parameters: $\lambda$ is the probability of being an inlier loop-closure; $u$ is the constant probability to uniformly sample a random loop-closure, which are difficult to set manually based on different datasets. Hence, we propose to estimate $\Theta$ based on the input data. More specifically, we learn $\Theta$ from the odometry constraints, since all odometry constraints are effectively inlier loop-closure constraints. 

The process to learn $\Theta$ is as follows. First, for each odometry constraint $X_{i,i+1}$, we denote its corresponding error term $m_{i,i+1} = e^{2A_{i,i+1}}$ (analogous to Eq.~(\ref{eq:lijin})), where
\begin{equation}\label{eq:aii+1}
    A_{i,i+1} = \frac{1}{\vert X_{i,i+1} \vert} \sum_{(\mathbf{p}, \mathbf{q}) \in X_{i,i+1}} {\ln (1 + \frac{\Vert T_i \mathbf{p} - T_{i+1} \mathbf{q} \Vert^2}{\sigma^2})}.
\end{equation}
Next, we compute the median error denoted as $\hat{m}$. Since we regard all odometry constraints as inlier loop-closure constraints, let
\begin{equation}\label{eq:phat}
    \frac{\Theta}{\Theta + \hat{m}} = \hat{p},
\end{equation}
where we set $\hat{p} = 90\%$, meaning that a loop-closure $Y_{ij}$ with a small error ($e^{2A_{ij}} < \hat{m}$) is very likely to be an inlier ($\mathcal{P}_{ij}^{in} > \hat{p}$). Finally, we solve for $\Theta$ using Eq. (\ref{eq:phat}).

\subsection{Maximization Step}
In the maximization step, we solve for $T$ that maximizes $Q_{EM} = Q_X + Q_Y$, where $Q_X$ and $Q_Y$ are shorthand notations defined in Eq.~(\ref{eq:q}).
These two terms are evaluated independently, and then optimized jointly.

\paragraph{Evaluate $Q_X$.} 
Assuming the odometry constraints in $X$ are i.i.d., the joint probability of all odometry constraints is given by
\begin{equation}\label{eq:x|t}
    p(X \vert T) = \prod_{i=1}^{N-1} p(X_{i,i+1} \vert T).
\end{equation}
Substituting the joint probability of the feature matches within each odometry constraint (Eq. (\ref{eq:xii+1})), we can rewrite $Q_X$ as
\begin{equation}\label{eq:qx}
    Q_X = - 2 \sum_{i=1}^{N-1} A_{i,i+1} + \text{const}.
\end{equation}

\paragraph{Evaluate $Q_Y$.}
Using the product rule, the joint probability of loop-closure constraints and their corresponding assignment variables can be written as $p(Y,Z \vert T) = p(Z) p (Y \vert Z, T)$. Plugging Eq. (\ref{eq:zij}), (\ref{eq:cij}) and (\ref{eq:yij|zout}) in, we have
\begin{align}\label{eq:yz|t}
    p(Y,Z \vert T) &= \prod_{i,j} (\lambda~C_{ij} )^{Z_{ij,in}}
    \big( (1-\lambda) u \big)^{Z_{ij,out}}.
\end{align}
We can rewrite $Q_Y$ as
\begin{equation}\label{eq:qy}
    Q_Y = \sum_{i,j} \mathcal{P}_{ij}^{in} \ln C_{ij} + \text{const},
\end{equation}
with the joint probability from Eq. (\ref{eq:yz|t}) and the posterior from Eq. (\ref{eq:lijin}), which can be further expanded to
\begin{equation}
    Q_Y = -2 \sum_{i,j} \mathcal{P}_{ij}^{in} A_{ij} + \text{const}.
\end{equation}

\paragraph{Maximize $Q_{EM}$.}
The maximization of $Q_X + Q_Y$ can be reformulated into a non-linear least-squares problem with the following objective function
\begin{align}
    \argmin_T 
    \sum_{i,j} \frac{\mathcal{P}_{ij}^{in}}{\vert Y_{ij} \vert}
    \sum_{(\mathbf{p}, \mathbf{q}) \in Y_{ij}} 
    &\ln (1 + \frac{\Vert T_i \mathbf{p} - T_{j} \mathbf{q} \Vert^2}{\sigma^2}) \\
    + \sum_{i=1}^{N-1} 
    \frac{1}{\vert X_{i,i+1} \vert}
    \sum_{(\mathbf{p}, \mathbf{q}) \in X_{i,i+1}} 
    &\ln (1 + \frac{\Vert T_i \mathbf{p} - T_{i+1} \mathbf{q} \Vert^2}{\sigma^2}), \notag
\end{align}
which can be easily optimized using the sparse Cholesky solver in Google Ceres \cite{ceres}. The computation complexity is cubic to the total number of feature matches.

\section{Generalization using EM}\label{sec:generalization}
In the previous section, we solved the problem when constraints are contaminated with outlier feature matches. In this section, we study a problem with an easier setting where correct loop-closure constraints contain no outlier feature matches. 
Recall that long-tail multivariate Cauchy distribution is used to gain robustness against outlier feature matches. 
We replace the multivariate Cauchy distribution with a multivariate Gaussian distribution for the easier problem without outlier feature matches, and show that our EM formulation degenerates to the formulation of a state-of-the-art approach for robust indoor reconstruction \cite{baseline}.
To avoid repetition, we only highlight the major differences to the previous section. Each analogous term is augmented with a superscript $G$ that stands for ``Gaussian''.

\paragraph{Odometry constraints.}
Replacing the multivariate Cauchy distribution in Eq. (\ref{eq:xii+1}) with a multivariate Gaussian distribution, we have
\begin{equation}\label{eq:xii+1,g}
    p^G(X_{i,i+1} \vert T)=\Big( \prod_{
    (\mathbf{p}, \mathbf{q}) \in X_{i,i+1}} \textit{Gauss}_{i,i+1}(\mathbf{p}, \mathbf{q})
     \Big)^{\frac{1}{\vert  X_{i,i+1} \vert}},
\end{equation}
where
\begin{equation}
    \textit{Gauss}_{i,i+1}(\mathbf{p}, \mathbf{q}) = \frac{\exp \big(-\frac{1}{2} d_{\Sigma}^2(T_i \mathbf{p}, T_{i+1} \mathbf{q}) \big)}{\sqrt{(2\pi)^3 \det\Sigma}},
\end{equation}
and $\Sigma$ and $d_{\Sigma}$ remain unchanged.

\paragraph{Loop-closure constraints.}
We note that the Bernoulli distribution in Eq. (\ref{eq:zij}) still holds, and the major changes start from Eq. (\ref{eq:cij}).
Using the multivariate Gaussian distribution, we have
\begin{equation}\label{eq:cij,g}
    G_{ij} := p^G(Y_{ij} \vert Z_{ij,in} = 1, T) =  (\sqrt{2\pi}\sigma)^{-3} e^{-\frac{B_{ij}}{2\sigma^2}},
\end{equation}
in which
\begin{equation}\label{eq:bij}
    B_{ij} = \frac{1}{\vert Y_{ij}\vert }
    \sum_{(\mathbf{p}, \mathbf{q}) \in Y_{ij}} 
    \Vert T_i \mathbf{p} - T_{j} \mathbf{q} \Vert^2,
\end{equation}
and $\vert Y_{ij}\vert$ is the number of feature matches.
We note that $B_{ij}$ is a sum-of-square errors that can lead to arithmetic overflow in the $e^{B_{ij}}$ term from the posterior of the latent variable $Z_{ij}$ (analogous to Eq. (\ref{eq:lijin})). In contrast, there is no arithmetic overflow in the $e^{A_{ij}}$ term from Eq. (\ref{eq:lijin}) since $A_{ij}$ from Eq. (\ref{eq:aij}) is a sum-of-log errors.
We propose to alleviate the arithmetic overflow problem by using a Pareto distribution that approximates $G_{ij}$ as
\begin{equation}\label{eq:pareto}
    G_{ij} \approx \frac{ x_0}{B_{ij}^2},
\end{equation}
where $x_0 > 0$ is a scale parameter.
For outlier loop-closures, the uniform distribution in Eq. (\ref{eq:yij|zout}) still holds.

\begin{table*}
\centering
 \begin{tabular}{l | l || c | c | c | c || c }
 \hline
 \multicolumn{2} {c||} {} & Living room 1 & Living room 2 & Office 1 & Office 2 & Average \\
 \hline
 \multirow{2}{*}{Before pruning} & Recall(\%) & 61.2 & 49.7 & 64.4 & 61.5 & 59.2 \\
 & Precision(\%) & 27.2 & 17.0 & 19.2 & 14.9 & 19.6 \\
 \hline
 Choi et al. \cite{baseline} & Recall(\%) & 57.6 & 49.7 & 63.3 & 60.7 & 57.8 \\
 after pruning & Precision(\%) & 95.1 & 97.4 & \ 98.3 & 100.0 & 97.7 \\
 \hline
 Ours (Sec. \ref{sec:generalization}) & Recall(\%) & {58.7} & 48.4 & {63.9} & {61.5} & {58.1} \\
 after pruning & Precision(\%) & {97.0} & 94.9 & 96.6 & 93.6 & 95.4 \\
 \hline
\end{tabular}
\vspace{3pt}
\caption{Results of robust optimization on the indoor dataset. Our method shows comparable result with the state-of-the-art.}
\label{tab:recallprecision}
\end{table*}

\begin{table*}
\centering
 \begin{tabular}{ l || c | c | c | c || c }
 \hline
  & Living room 1 & Living room 2 & Office 1 & Office 2 & Average \\
 \hline
 Whelan et al. \cite{kintinuous} & 0.22 & 0.14 & 0.13 & 0.13 & 0.16 \\
 \hline
 Kerl et al. \cite{dvoslam} & 0.21 & {0.06} & 0.11 & 0.10 & 0.12 \\
 \hline
 SUN3D \cite{sun3d} & 0.09 & 0.07 & 0.13 & 0.09 & 0.10 \\
 \hline
 Choi et al. \cite{baseline} & {0.04} & 0.07 & {0.03} & {0.04} & {0.05} \\
 \hline 
 Ours (Sec. \ref{sec:generalization}) & 0.06 & 0.09 & 0.05 & {0.04} & 0.06 \\
 \hline\hline
 GT Trajectory & 0.04 & 0.04 & 0.03 & 0.03 & 0.04 \\
 \hline
\end{tabular}
\vspace{3pt}
\caption{Reconstruction accuracy on the indoor dataset. The entries are the mean distances of each model to its respective ground-truth surface (in meters). Our proposed method shows comparable result with the state-of-the-art and outperforms the rest.}
\label{tab:reconstructionaccuray}
\end{table*}

\paragraph{Expectation step.}
Using the approximation of $G_{ij}$ in Eq. (\ref{eq:pareto}), the posterior from Eq. (\ref{eq:lijin}) becomes
\begin{equation}
    {\mathcal{P}_{ij}^{in}}^{G} \approx \frac {\Theta^G}{\Theta^G + B_{ij}^2},
\end{equation}
where 
\begin{equation}
    \Theta^G = \frac{x_0 \lambda}{(1 - \lambda) u} .
    \label{eq:G_ij}
\end{equation}
It becomes apparent in $\mathcal{P}_{ij}^{{in}^{G}}$ that the arithmetic overflow problem is alleviated by the replacement of $e^{B_{ij}}$ with $B_{ij}^2$.
In the previous section, $\Theta$ in Eq.~(\ref{eq:phat}) is learned from the median error $\hat{m}$ of all the error terms $m_{i,i+1} = e^{2A_{i,i+1}}$ in the odometry constraints. Unfortunately, the median error $\hat{m}^G$ from $m_{i,i+1}^G = B_{i,i+1}^2$ becomes uninformative because we assume no outlier feature matches, \ie, $\hat{m}^G \rightarrow 0$ since $m_{i,i+1}^G = B_{i,i+1}^2 \rightarrow 0$.
Despite the absence of outlier feature matches, $\Vert T_i \mathbf{p} - T_j \mathbf{q} \Vert$ is upper bounded by some threshold, $\epsilon$. Hence, the mean error term can be directly estimated from Eq. (\ref{eq:bij}) as $\hat{m}^G = {\epsilon^2}$. Subsequently, let
\begin{equation}
    \frac{\Theta^G}{\Theta^G + \hat{m}^G} = \hat p,
\end{equation}
where we set $\hat p = 90\%$ and solve for $\Theta^G$. We set $\epsilon = 0.05$m for our experiments on the indoor dataset (see next section) based on the typical magnitude of sensor noise.

\paragraph{Maximization step.}
Finally, we reformulate the maximization problem as a non-linear least-squares problem with the following objective function
\begin{align} \label{eq:nonlinearLeastsqs}
    \argmin_T 
    \sum_{i,j} 
    \frac{{\mathcal{P}_{ij}^{in}}^G}{\vert Y_{ij} \vert }
    \sum_{(\mathbf{p}, \mathbf{q}) \in Y_{ij}} 
    & \Vert T_i \mathbf{p} - T_{j} \mathbf{q} \Vert^2 \\
    + \sum_{i=1}^{N-1} 
    \frac{1}{\vert X_{i, i+1} \vert }
    \sum_{(\mathbf{p}, \mathbf{q}) \in X_{i,i+1}}
    & \Vert T_i \mathbf{p} - T_{i+1} \mathbf{q} \Vert^2, \notag
\end{align}
which is similar to the formulation in \cite{baseline} with two minor differences. First, we average the square errors over the number of feature matches but \cite{baseline} does not. Second, we estimate the posterior ${\mathcal{P}_{ij}^{in}}^G$ by iterating between the Expectation and Maximization steps but \cite{baseline} estimates it using line processes \cite{lineprocess}. It is important to note that Eq. (\ref{eq:nonlinearLeastsqs}) is derived from the original Gaussian formulation in Eq. (\ref{eq:cij,g}) instead of the Pareto approximation in Eq. (\ref{eq:pareto}).

\section{Evaluation}\label{sec:eval}
We use the experimental results from two datasets for the comparison between
our approach and the state-of-the-art approach \cite{baseline}. The first dataset is from small-scale indoor scenes with no outlier feature matches in the odometry and inlier loop-closure constraints, and the 
second dataset is from large-scale outdoor scenes with outlier feature matches. Our Gaussian-Uniform EM (Sec.~\ref{sec:generalization}) and Cauchy-Uniform EM (Sec.~\ref{sec:EM}) are evaluated on the small-scale indoor and large-scale outdoor datasets, respectively.


\subsection{Small-Scale Indoor Scenes}


The ``Augmented ICL-NUIM Dataset" provided and augmented by \cite{handa2014benchmark} and \cite{baseline}, respectively, is used as the small-scale indoor dataset. This dataset is generated from synthetic indoor environments and includes two models: a living room and an office. There are two RGB-D image sequences for each model, resulting in a total of four test cases. To ensure fair comparison, we follow the same evaluation criteria and experimental settings as \cite{baseline}.

\paragraph{Results.}
Tab. \ref{tab:recallprecision} shows the comparison of the average recall and precision of the loop-closures on (1) before pruning, (2) \cite{baseline} after pruning and (3) our method after pruning. Here, ``before pruning" refers the loop-closures from the loop-closure detection, and ``after pruning" refers to the inlier loop-closures after robust optimization. It can be seen that the average precision and recall of our method is comparable to \cite{baseline}. This is an expected result since we 
showed in Sec. \ref{sec:generalization} that our method degenerates to the method in \cite{baseline} with minor differences in the absence of outlier feature matches.
We further evaluate the reconstruction accuracy of the final model using the error metric proposed in \cite{handa2014benchmark}, \ie, the mean distance of the reconstructed surfaces to the ground truth surfaces. Tab. \ref{tab:reconstructionaccuray} shows the comparison of the reconstruction accuracy of our method to other existing approaches. In addition, as suggested in \cite{baseline}, the reconstruction accuracy of the model obtained from fusing the input depth images with the ground truth trajectory (denoted as GT Trajectory in Tab. \ref{tab:reconstructionaccuray}) is reported for reference. 
As expected, our method shows comparable result with the state-of-the-art on the indoor dataset.

\begin{figure*}
    \centering
    \includegraphics[width=1\textwidth]{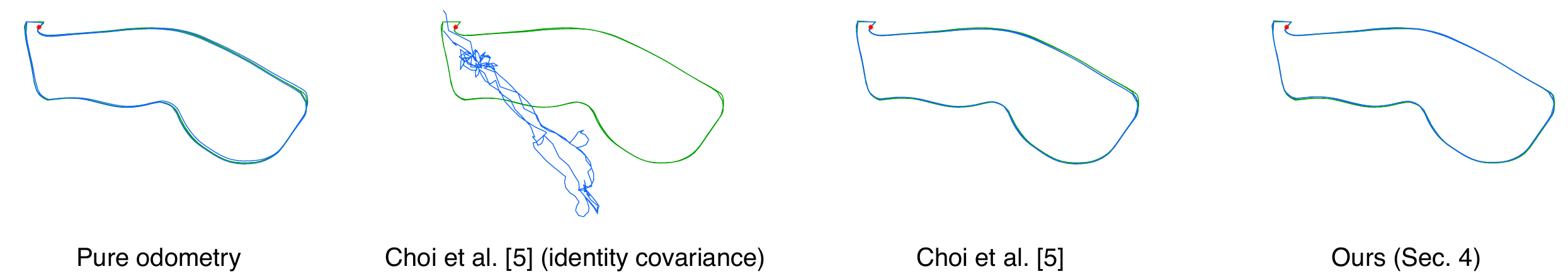}
    \caption{Trajectories on the 1km route. Each trajectory (blue) is overlaid with the GPS/INS trajectory (green). Red asterisk indicates the position of the 1st fragment pose.}
    \label{fig:short-trajectory}
\end{figure*}

\begin{figure*}
    \centering
    \includegraphics[width=1\textwidth]{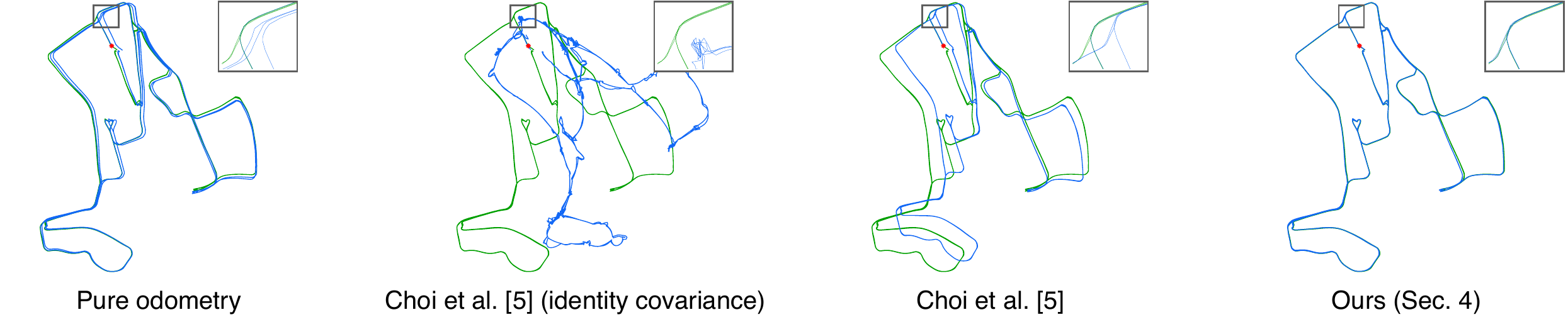}
    \caption{Trajectories on the city-scale route. Each trajectory (blue) is overlaid with the GPS/INS trajectory (green). A zoomed-in region is shown on the top right corner for each trajectory. Red asterisk indicates the position of the 1st fragment pose.}
    \label{fig:long-trajectory}
\end{figure*}

\subsection{Large-Scale Outdoor Scenes}
The large-scale outdoor dataset is based on the ``Oxford Robotcar Dataset" \cite{oxford}. It consists of 3D point clouds captured with a LiDAR sensor mounted on a car that repeatedly drives through Oxford, UK, at different times over a year. 
We select two different driving routes from the dataset, a short route (about 1km) and a long route (city-scale).
Furthermore, we take two traversals at different times for each route, resulting four traversals in total. Unlike the synthetic indoor dataset, there is no ground truth of the surface geometry. We evaluate the trajectory accuracy against the GPS/INS readings as an indirect measurement of reconstruction accuracy. We prepare the dataset as follows:

\begin{itemize}[leftmargin=*]
    \item \textbf{Point cloud fragments.} We integrate the push-broom 2D LiDAR scans and their corresponding INS readings into the 3D point clouds. We segment the data into fragments with 30m radius for every 10m interval. Each fragment is then downsampled using a VoxelGrid filter with a grid size of 0.2m. 
242 and 1770 fragments are constructed for the 1km route and the city-scale route, respectively.
   \item \textbf{Odometry trajectory.}
The odometry trajectory is disconnected due to discontinuous INS data
since we are combining two traversals.
We simulate the odometry trajectory via geometric registrations between consecutive point cloud fragments, and manually identify one linkage transformation between the two traversals. We also check the entire odometry trajectory to ensure that there are no remaining erroneous transformations. The resulting odometry trajectory is used to initialize the fragment poses, $T$. 
   \item \textbf{Odometry constraints.}
For every two consecutive frames along the odometry trajectory, we perform point cloud registration as described in Sec. \ref{sec:overview}. Specifically, we extract 1024 features for each fragment, and collect the top 200 feature matches to form an odometry constraint. Note that the feature matches are selected without additional geometric verification, and it can contain outliers. 
241 and 1769 odometry constraints are constructed for the 1km route and the city-scale route, respectively.
   \item \textbf{Loop-closure constraints.}
We perform loop-closure detection as described in Sec. \ref{sec:overview}. We take every 5th fragment along the trajectory as a keyframe fragment; loop-closures are detected among the selected keyframe fragments. 
For the 1km route, we find the top 5 loop-closures for each keyframe fragment and then remove the duplicates. For the city-scale route, we find the top 10 loop-closures for each keyframe fragment and then remove the duplicates.
171 and 1438 loop-closure constraints are constructed for the 1km and city-scale route, respectively. The outlier loop-closure ratio is more than 80\% for both routes.
\end{itemize}

\begin{figure*}
    \centering
    \includegraphics[width=0.95\textwidth]{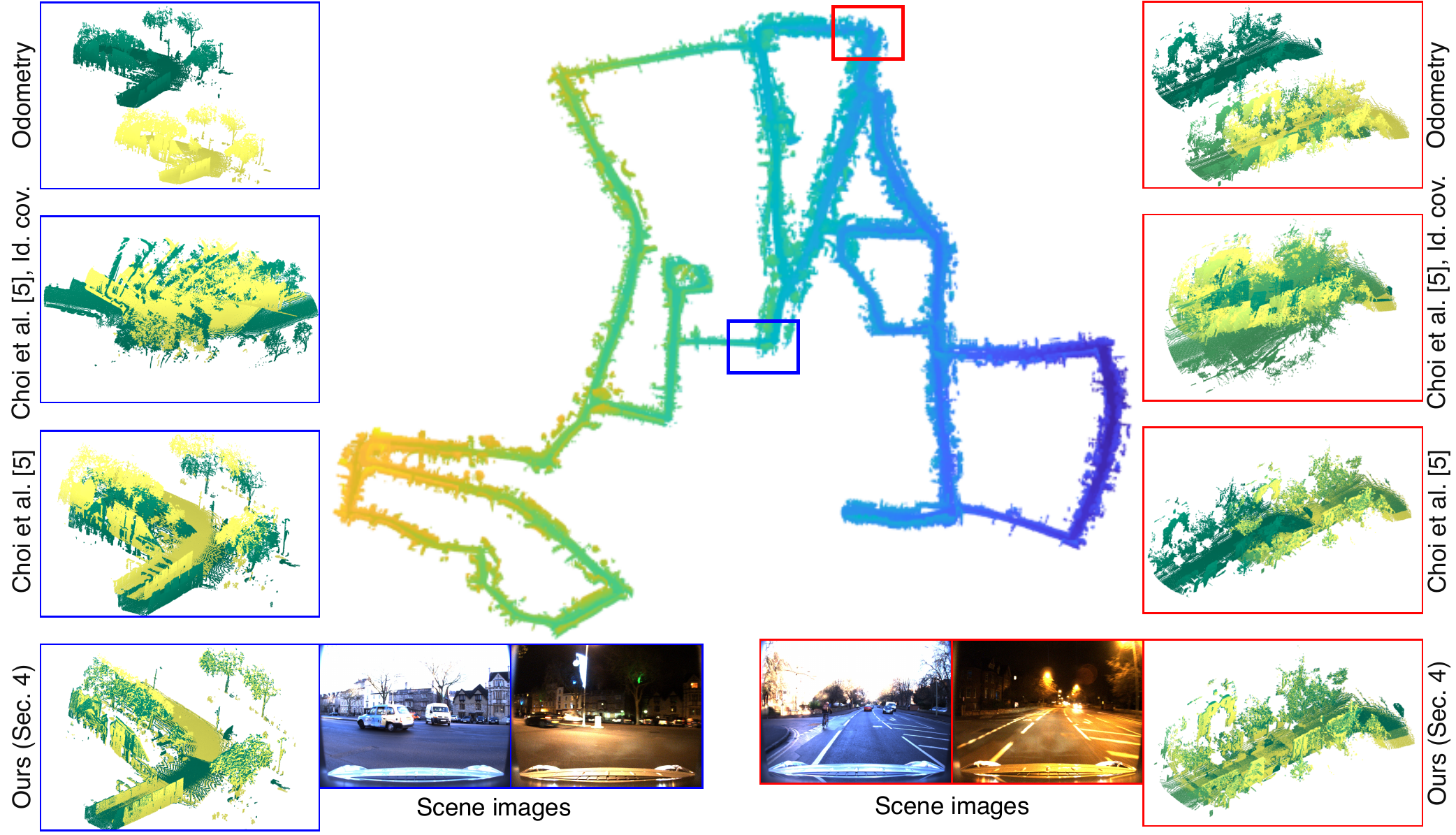}
    \caption{Reconstruction results on the city-scale route traversed in day and night. The two columns of zoomed-in point clouds are reconstructed based on different trajectories. }
    \label{fig:long-link}
\end{figure*}

\begin{table}
\centering
 \begin{tabular}{ c || c | c }
 \hline
  & 1km & City-scale  \\
 \hline
 \multirow{2}{*}{Odometry} & \multirow{2}{*}{1.85} &  \multirow{2}{*}{11.81}  \\
 & &\\
 \hline
 Choi et al. \cite{baseline}  & \multirow{2}{*}{123.24} & \multirow{2}{*}{207.93} \\
 (identity covariance) & &\\
 \hline
 \multirow{2}{*}{Choi et al. \cite{baseline}} & \multirow{2}{*}{1.97} & \multirow{2}{*}{50.92}  \\
 & &\\
 \hline
 \multirow{2}{*}{Ours (Sec. \ref{sec:EM})} & \multirow{2}{*}{\textbf{1.34}} &  \multirow{2}{*}{\textbf{2.45}}  \\
 & &\\
 \hline
\end{tabular}
\vspace{3pt}
\caption{Reconstruction accuracy on outdoor dataset. Each entry is the mean distance of the estimated poses to the GPS/INS ground truth (in meters).}
\label{tab:outdoorresults}
\end{table}

\vspace{-10pt}
\paragraph{Baseline Methods.}
We compare the effectiveness of our approach with two baseline methods based on \cite{baseline}: a stronger and a weaker baseline. The stronger baseline encodes uncertainty information of the feature matches between two fragments into a covariance matrix. The feature matches used to construct the covariance matrix are those within 1m apart after geometric registration. Refer to \cite{baseline} for the more details on the covariance matrix.
The covariance matrix of the weaker baseline is set to identity, \ie, no uncertainty information on the feature matches. The relative poses between the point cloud fragments computed from ICP are used as the odometry and loop-closure constraints in the baseline methods. 


\paragraph{Results.}
Tab. \ref{tab:outdoorresults} summarizes the mean distances of the estimated poses to the GPS/INS trajectory as an indirect measure of the reconstruction accuracy on the 1km and city-scale outdoor datasets. 
Fig. \ref{fig:short-trajectory} and \ref{fig:long-trajectory} show the plots of the trajectories.
We align the first five fragment poses with the GPS/INS trajectory, error measurements start after the 5th fragment pose. The results show that the accuracy increases when more information about the feature matches is considered in the optimization process. 
We can see from Tab. \ref{tab:outdoorresults}, and Fig. \ref{fig:short-trajectory} and \ref{fig:long-trajectory} that
the weaker baseline (\cite{baseline} with uninformative identity covariance) without information of the feature matches gives the worst performance. The stronger baseline (\cite{baseline} with informative covariance matrix) that encodes information of feature matches using the covariance matrix shows better performance.
In contrast, our method that directly takes feature matches as the odometry and loop-closure constraints outperforms the two baselines. 
Furthermore, Fig. \ref{fig:short-link} and \ref{fig:long-link} show reconstruction results for qualitative evaluation. It can be seen from the bottom left and right plots in Fig. \ref{fig:long-link} that our method produces the sharpest reconstructions of the 3D point clouds.


\vspace{-5pt}
\section{Conclusion}
In this paper, we proposed a probabilistic approach for robust point cloud reconstruction of large-scale outdoor scenes. 
Our approach leverages on a Cauchy-Uniform mixture model to simultaneously suppress outlier feature matches and loop-closures. Moreover, 
we showed that by using a Gaussian-Uniform  mixture  model,  our  approach  degenerates  to  the formulation of a state-of-the-art approach for robust indoor reconstruction. We verified our proposed methods on both indoor and outdoor benchmark datasets.


\section*{Acknowledgement}
This work is supported in part by a Singapore MOE Tier 1 grant R-252-000-A65-114.

{\small
\bibliographystyle{ieee}
\bibliography{egbib}
}
\end{document}